\documentclass[conference]{IEEEtran}
\IEEEoverridecommandlockouts
\usepackage{cite}
\usepackage{amsmath,amssymb,amsfonts}
\usepackage{algorithmic}
\usepackage{graphicx}
\usepackage{textcomp}
\usepackage{xcolor}

\usepackage{multirow} 
\usepackage{multicol}

\def\BibTeX{{\rm B\kern-.05em{\sc i\kern-.025em b}\kern-.08em
    T\kern-.1667em\lower.7ex\hbox{E}\kern-.125emX}}
\begin{document}

\title{Contrastive Semantic Similarity Learning for Image Captioning Evaluation with Intrinsic Auto-encoder \\
{\footnotesize \textsuperscript{}
}  % add head notes here
\thanks{
} % foot notes
}

%\author{\IEEEauthorblockN{1\textsuperscript{st} Given %Name Surname}
%\IEEEauthorblockA{\textit{dept. name of organization (of %Aff.)} \\
%\and

\author{\IEEEauthorblockN{Chao Zeng}
	\IEEEauthorblockA{\textit{City University of Hong Kong} \\
		\textit{chao.zeng@my.cityu.edu.hk}\\
		}
	\and
	\IEEEauthorblockN{Tiesong Zhao}
	\IEEEauthorblockA{\textit{Fuzhou University} \\
		\textit{t.zhao@fzu.edu.cn}\\
		}
	\and
	\IEEEauthorblockN{Sam Kwong}
	\IEEEauthorblockA{\textit{City University of Hong Kong} \\
		\textit{cssamk@cityu.edu.hk}\\
	}

}

\maketitle

\begin{abstract}
	
Automatically evaluating the quality of image captions can be very challenging since human language is quite flexible that there can be various expressions for the same meaning. Most of the current captioning metrics rely on token level matching between candidate caption and the ground truth label sentences. It usually neglects the sentence-level information. Motivated by the auto-encoder mechanism and contrastive representation learning advances, we propose a learning-based metric for image captioning, which we call Intrinsic Image Captioning Evaluation($I^2CE$). We develop three progressive model structures to learn the sentence level representations--single branch model, dual branches model, and triple branches model. Our empirical tests show that $I^2CE$ trained with dual branches structure achieves better consistency with human judgments to contemporary image captioning evaluation metrics. Furthermore, We select several state-of-the-art image captioning models and test their performances on the MS COCO dataset concerning both contemporary metrics and the proposed $I^2CE$. Experiment results show that our proposed method can align well with the scores generated from other contemporary metrics. On this concern, the proposed metric could serve as a novel indicator of the intrinsic information between captions, which may be complementary to the existing ones.

\end{abstract}


\section{Introduction}
\begin{figure}[htbp]
	\centerline{\includegraphics[width=9.5cm]{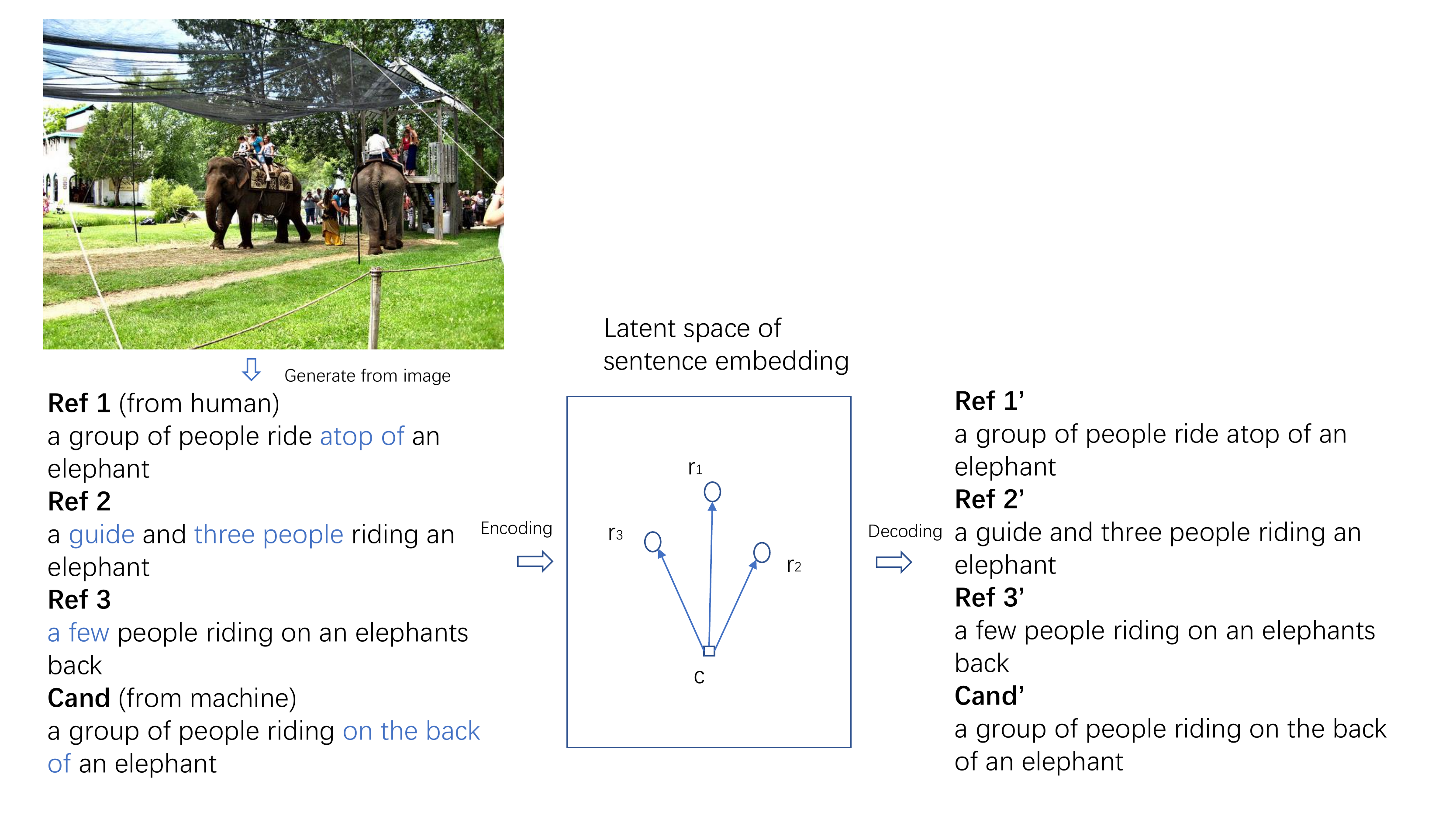}}
	\caption{Intrinsic variance in ground truths and candidate. }
	\label{fig}
\end{figure}
Generating proper descriptions from images to language expression has gained much attention from computer vision and natural language processing researchers. Moreover, since neural networks prosper, the research community has witnessed the proliferation of various neural image captioning models. However, The evaluation metrics for image captioning seem to remain unchanged for years, such as BLEU\cite{papineni2002bleu}, METEOR\cite{banerjee2005meteor}, ROUGE\cite{lin2004rouge}, CIDER\cite{vedantam2015cider} and SPICE \cite{anderson2016spice}, all aiming to calculate an alignment between candidate captions and ground truth sentences. 
There are some key challenges faced with the automatic evaluation metrics for image captioning. Firstly the contemporary metrics are struggling with the problem of deviating from human judgments. Thus metrics based on token overlap would miss the semantic similarity among sentences. For one thing, there can be various expressions pointing to the same meaning. For another, there might be altering meanings of the same word, which are all difficult to cover for the traditional metrics. Secondly, there can be various blind spots for rule-based metrics. For example, the SPICE metric excels in capturing the visual information by scene graphs but not as good for sentence structural information at the sentence level. As far as we know, sentence level embedding is currently not as explored as token level embedding.
Generally, the contemporary metrics provide different approaches to calculating the similarity degree between the generated captions and the ground truth captions in the dataset. In the captioning dataset, there are generally five ground truth sentences. As shown in Fig. 1, there are intrinsic variances in the annotations with different concerns or expressions on phrases.

To address the challenges mentioned above, we propose a learning-based metric that captures intrinsic information entailed among different sentences. We use auto-encoder to recover the input sentence from itself. In an auto-encoder, the input would be first transformed into a vector representation with an encoder, followed by a decoder to generate the original input. Here the supervision information is from the input data itself. Inspired by this, we assume that if the intrinsic representation can recover the original data, then the intrinsic representations can be used to calculate the similarity among samples. We carry out the proposed Intrinsic Image Captioning Evaluation ($\rm{I^2CE}$) based on the above idea. 
We use MSCOCO label sentences to train the sentence auto-encoder. To learn a more semantic distance-aware model, we regularize the original sentence reconstruction loss with a triplet embedding loss term. In this way, the model will also learn to push negative sentence pairs far away and pull similar sentences closes in the semantic embedding space. Furthermore, we select several state-of-the-art image captioning models to test their performances with respect to conventional metrics and our $\rm{I^2CE}$. Experiments results show the effectiveness and robustness towards captions of different qualities.The main contributions of this paper include:

\textbf{·} We propose the Intrinsic Image Captioning Evaluation  ($\rm{I^2CE}$) metric, a self-supervised learning method based on auto-encoding mechanism and contrastive semantic learning. The existing metrics typically use token level matching, which may lose sentence level information. On contrary, our proposed method generate sentence level embedding to calculate semantic similarity.

\textbf{·} We demonstrate how to utilise auto-encoding mechanism to learn sentence level embedding for semantic representation. Moreover, to make the learned representation more distance-aware, we explicitly add a semantic loss term in the overall training objective. And by forming training corpus in sentence pairs or triplets, we develop two progressive model structures based on the original single-branch structure.

\textbf{·} We perform an empirical study on the human judgments correlation for both the proposed method and the contemporary adopted metrics. The results show that $\rm{I^2CE}$ has a higher consistency with human judgments. In addition, we test performances of various state-of-the-art image captioning models on the MS COCO dataset with both contemporary metrics and $\rm{I^2CE}$. More over, our intuitive results shows that $\rm{I^2CE}$ has dynamic and highly semantic related properties on scoring for testing captions.

The remainder of this paper is organized as follows. Section 2 includes related work, and in Section 3 we proposed our $\rm{I^2CE}$ method. The dataset,experiments and intuitive results are then described in Section 4. And we conclude the paper with Section 5.

\section{Related Work}

\subsection{Image Captioning}
Early image captioning models generate captions by translating detected concept words to sentences with a template\cite{farhadi2010speaking}. In recent years the encoder-decoder framework based Neural Image Captioning model was proven effective in this image to text translation problem\cite{vinyals2015show}. Later as the attention mechanism introduced, the captioning performance gets improved with a great margin by attention models like spatial attention\cite{xu2015show, anderson2018bottom}, semantic attention\cite{you2016image}, adaptive attention\cite{lu2017knowing},etc. These models share an encoder-decoder framework. Generally, the captioning model uses a CNN encoder to get the image representations and use an RNN model as a decoder to decode from the image features. More recently, transformers
\cite{vaswani2017attention, devlin2018bert} show excellent performance in natural language processing tasks, in which the multi-head self-attention plays a key role in modeling the contextual relationship among tokens. Marcella et al.\cite{cornia2020meshed} present a Meshed Transformer with Memory for image captioning and show state-of-the-art performances. Huang et al.\cite{huang2019attention} build attention on attention framework and also achieve great performances. Guo et al. use the transformer differently and build a non-auto regressive image captioning model with multi-agents optimization\cite{guo2020non}
. In summary,  the image captioning system contains three parts: the visual encoder, the language decoder, and the visual-textual interactive part.

\subsection{Caption Evaluation Metrics}
$BLEU$ is a precision-based n-gram matching metric which was originally designed for machine translation evaluation\cite{papineni2002bleu}. The BLEU calculates the proportion of how much of the generated N-grams in a candidate sentence got matched with the ground truth references.  By combining the contributions of different N-grams(N equals to 1,2,3, or 4) it then takes average to get the final score of the candidate sentence with the short length punishing term attached. 

The $ROUGE$ metric concerns on balance between the recall and precision, while BLEU only considers the precision of the generated candidate. It was originally proposed for text summary tasks. There are different branches of the $ROUGE$ algorithm, among which $ROUGE-L$ uses the longest common sub-sequence. And $ROUGE-L$ is commonly used for image captioning evaluation.

$METEOR$ is designed for machine translation. Based on the F measure of the previous ROUGE metric,it introduces modifier of the chunk punishment term. Automatically this algorithm will find the non-intersecting matching chunk pairs between the candidate and the reference. 

$CIDER$ is a metric specially designed for image captioning evaluation. The core mechanism embedded in CIDER is the tf-idf weighting term, which has a wide range of applications in vector semantics. It represents a sentence as an n-gram vector regularized with tf-idf(Term Frequency-Inverse Document Frequency) term. The cosine distance between the candidate and the ground truth is then used as the semantic relevance.

The $SPICE$ metric uses the graph representation of sentences.
The matching process is carried out through the comparison between two tuple sets of textual and image graphs. However, the performance of this metric relies much on the parsing accuracy. It prefers longer sentences and tends to lose structural information of sentences.

$WMD$ metric takes the similarity problem between candidate and reference captions as a special instance of the Earth Mover’s Distance, which is a well-known transportation linear optimization problem\cite{kusner2015word}. The assumption is that the more similar the two captions, the less cost it would take to transfer one to the other.

Also, there are some recently proposed metrics for image captioning evaluation. On the concern on diversity in image caption generation, Qingzhong  et al. proposed a diversity indicator as a metric\cite{wang2019describing}. The main idea is that for a specific image, sample from a captioning model to generate multiple different captions and then conduct a k-SVD decomposition analysis on these captions. If the singularity values are balanced well, then the model is good at diversity property. On the concern on distinctiveness, Ruotian et al. introduce an image retrieval auxiliary image caption generation model\cite{luo2018discriminability}. In the experiment, the margin loss term is introduced for generating more distinctive captions. For measuring the distinctiveness of the generated captions, they utilize an image retrieval model to provide with feedbacks and relay this extra signal back to the caption generation model. Jiuniu et al. present a distinctiveness-oriented caption metric\cite{wang2020compare}. By building similar caption sets and giving more weights on those distinct words the caption model will learn to emphasize more on the distinctive parts of the image.

BERTScore\cite{zhang2019bertscore} is the latest proposed sentence similarity metric based on token level embeddings learned from pretrained models. Yi et al. \cite{yi2020improving}propose a metric based on BERTScore to deal with the over-penalization problem of evaluation. They use a method to combine all five ground truths to a compact token set with redundancies removed and use this combined set as ground truth to compare with the candidate caption.
Xie et al. present a metric that considers grammaticality, accuracy and diversity \cite{xie2019going}. However, the evaluation is conducted on synthetic datasets.

To combine the visual informantion when evaluation Jiang et al. propose to use both regions grounding vectors and region attention weights distribution to calculate caption quality\cite{jiang2019tiger}. Based on joint visual textual embedding space Jiang et al. propose to compute a context vector of token embeddings attending to visual regions \cite{jiang2019reo}. Then the context vector of candidate caption and ground truths can be used to calculate the vector similarity as relevance. Some recent works consider evaluating the image descriptions with the actual image contents instead of the paired annotations \cite{madhyastha2019vifidel, agarwal2020egoshots}

Cui et al.\cite{cui2018learning} propose a caption metric model based on binary classifier. The model is trained by two kinds of captions: the human drafted ones and the machine-generated ones togather with data augamentation, which are labeled one or zero. Finally, the distinguishing model would give a score which is between zero and one, indicating how likely a caption is generated by a human. This probability score is taken as relativeness to human judgments. However, this metric has no explicit semantic distance mechanism embedded. On contrary, in $\rm{I^2CE}$ we explicily introduce a semantic distance loss term as a regularization term. 

Generally, most of the above metrics deal with semantic meanings at the token level whether rule based or learning based. And some of above works propose to include the visual features for caption evaluation. However sentence level embedding seems to be less explored for semantic evaluation. In this work we aim to learn a distance-aware semantic embedding at sentence level for image captioning evaluation task.

\subsection{Sentence Embedding}
Sentence similarity has been gaining attention from the natural language processing research community. The Siamese Manhattan LSTM \cite{mueller2016siamese} form this problem in a fully supervised way and employ two identical LSTM to learn the two input sentences according to human-annotated sentence similarity scores. SentenceBert \cite{reimers2019sentence} uses the powerful language model Bert as the backbone and a Siamese network structure to learn the similarity model.
These previous works need human-annotated scores as the supervision signal to train the similarity model. However, human annotations can be very expensive and with low efficiency. Motivated by the auto-encoding mechanism, we propose to learn a sentence representation in a self-supervised manner without the need for annotation data such as human-generated similarity scores.

In this paper, we aim to obtain a self-supervised learning-based image captioning evaluation metric, which can generate scores for captions without the usage of human-labeled scores. To achieve this goal, we train an auto-encoder to extract the gist embedded in the caption so that captions with similar semantic meaning would be mapped to neighboring areas, and captions with different meanings would be kept apart on a vector distance metric like cosine similarity. Our work mainly contains two components, the self-supervised autoencoder to learn the representation of caption and several strategies for training sentence embeddings to make the model able to distinguish between sentences with different meanings.

The framework of the proposed method i2ce is outlined in Figures 1,2, and 3. We develop three different settings of the proposed method, i.e., One branch approach, dual branches approach, and triple branches approach. The weights of encoders and decoders of each branch are shared, respectively. 

As shown in Fig. 1, we employ an auto-encoder for the word vector fusion. The left half is the encoder for extracting the meaning of the sentence, which is represented by the last hidden state vector of the encoder, which we name Intrinsic Vector. The right half is the decoder. It is only employed in the training stage to make learning the intrinsic encoder possible. 

We think that if the auto-encoder can reconstruct the input sentence itself after an abundant training process, then the intrinsic vector between the encoder and decoder can be taken as the meaning of the input sentence.

After training, we utilize the encoder to infer sentence representation as intrinsic vector for calculating the similarity between the candidate and reference captions.

\begin{figure}[htbp]
	\centerline{\includegraphics[width=9.5cm]{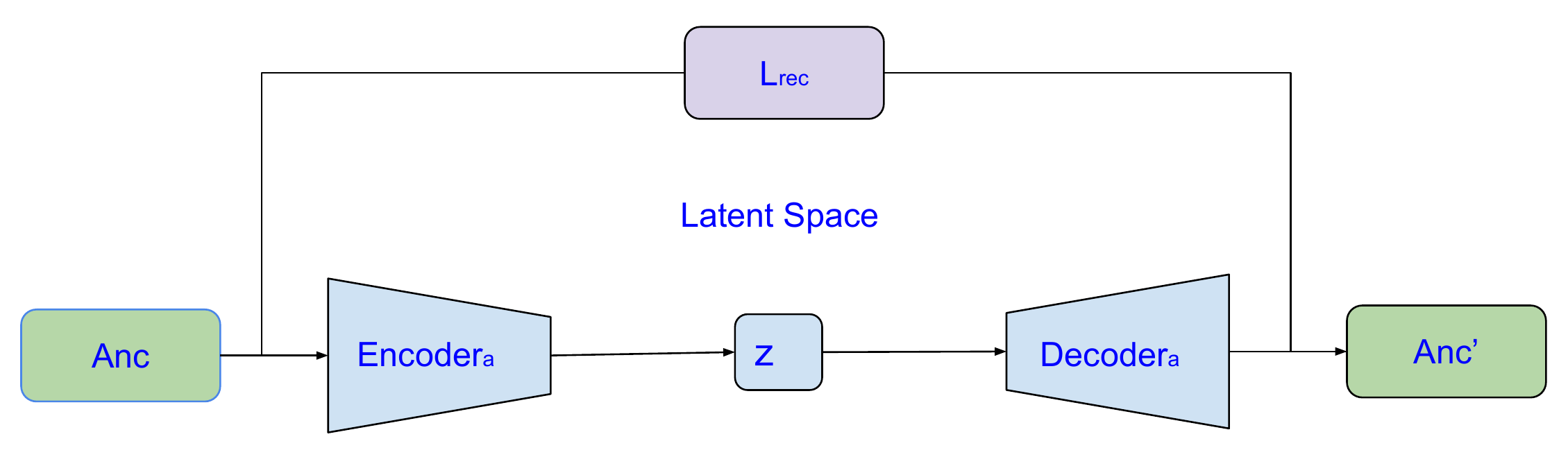}}
	\caption{$\rm{I^2CE}$ of single branch. The model mainly contains an textual encoder and decoder. $L_{rec}$ means sentence reconstruction loss. Anc stands for the input sentence and Anc' is the reconstructed sentence from decoder with the latent code z.}
	\label{fig}
\end{figure}

\subsection{Sentence Auto-Encoder with GRU}\label{AA}
In this work we use GRU\cite{chung2014empirical} to read in word level embeddings from an embedding layer and employs its final hidden state as a vector representation for each sentence. And the decoder receives the latent representation to generate the original input sentence. The machanism of GRU is shown in the following equations:
\begin{equation}
{{\rm{r}}_t} = \sigma ({W_r}{x_t} + {U_r}{h_{t - 1}} + {b_r})
\end{equation}
\begin{equation}
{{\rm{z}}_t} = \sigma ({W_z}{x_t} + {U_z}{h_{t - 1}} + {b_z})
\end{equation}
\begin{equation}
h_t^* = {\mathop{\rm Tanh}\nolimits} ({W_h}{x_t} + {U_h}({r_t} \odot h) + {b_h})
\end{equation}
\begin{equation}
{h_t} = (1 - {z_t}){h_{t - 1}} + {z_t}h_t^*.
\end{equation}
Here $r_t$ and $z_t$ are the reset gate and refresh gate, respectively. $x_t$ is the input embeddings, and $h_t$ is the hidden state of time step t. $z_t$ is a weight to measure how much information we need to keep for the current information flow $h_t^*$. $W$ and $U$ are matrices of a linear layer, $b$ is a bias of the linear layer. Delta and the tanh are activation functions. Compared to LSTM, which has three gates, GRU is a more lightweight model with fewer parameters. For the middle-level scale of training data, GRU can reach comparable or even better performance comparing to LSTM\cite{chung2014empirical}. In this study, we use GRU to both the encoder and decoder side of the auto-encoder model.

For the one branch approach, we use the sentence reconstruction loss to guide the training process. Reconstruction loss is to measure the loss between generated sentence from the decoder and the original input sentence. Here we adopt the NLL loss to calculate the word level loss between tokens and add up to the overall loss of the whole sentence for reconstruction. The reconstruction loss is shown as
\begin{equation}
{\rm{L}_{\rm{rec}}} = \sum\limits_{n = 1}^N {\sum\limits_{t = 1}^T {\rm{NLL}({y_t},{w_t})} }, 
\end{equation}
where $y_t$ is the ground truth word and the $w_t$ is the predicted word at time step $t$. $N$ and $T$ are the batch size and the maximum length of sequences. We use teacher forcing when at the training phase.

\begin{figure}[htbp]
	\centerline{\includegraphics[width=9.5cm]{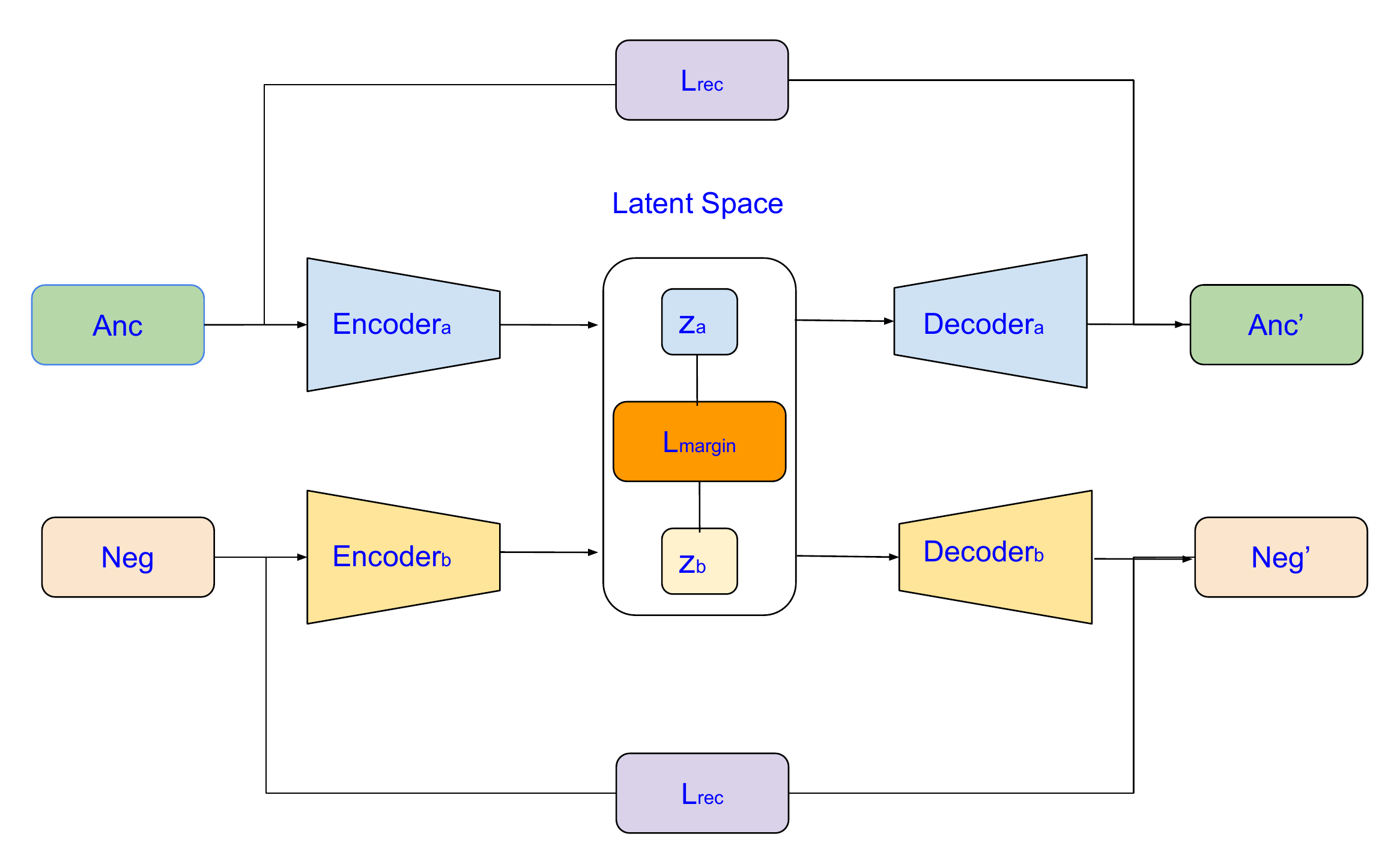}}
	\caption{$\rm{I^2CE}$ of dual branches.In latent space the margin loss between anchor(Anc) and negative(Neg) sentences is introduced. Weights are shared between the two encoders or decoders.}
	\label{fig}
\end{figure}

\subsection{Training with dual branches framework}
In the previous section, we could train an auto-encoder with the sentence reconstruction loss. The auto-encoder would automatically extract the most meaningful information embedding in the sentence to reconstruct itself and thus learn a latent representation of sentences. However, this training objective does not cover any distance-aware mechanism. To make the learned embedding more aware of the semantic distance, we introduce the margin loss between negtive sentences in this section. The margin loss is formed in the following equation
\begin{equation}
{{\rm{L}}_{\rm{margin}}}(a,a') = \mathop {{\mathop{\rm m}\nolimits} ean}\limits_{a'} {[m - d(f(a),f(a'))]_ + },
\end{equation}
where the plus label stands for taking the value in the bracelet when it is a positive value and zero when opposite. The $m$ here is a margin, and the $d$ is a function to measure semantic distance. In this work, we use Cosine distance between the sentence embeddings. The $a’$ stands for the randomly selected caption in the batch. Generally, in a batch, the sentences are all from the ground truth of different paired images. Thus, the sentences in a batch should bear different meanings, and we collect every pair within a batch and take the mean average of semantic distance as the contrast loss. In the experiment, we also explored maximum operation, which means only keep the hardest negative pair for the margin loss. We found that mean strategy shows better performance with regard to human judgments.

We can see that the marginal objective is actually forming the negative pairs within a batch to make the learned embedding distance aware in the semantic space. The distance-aware mechanism is explicitly embedded in the training process as the margin loss. This margin loss can be taken as a regularization mechanism on the original learned intrinsic embedding. With the guide of this distance-aware objective the model will learn both intrinsic and semantic distance aware representation of sentences.

\begin{figure}[htbp]
	\centerline{\includegraphics[width=9.5cm]{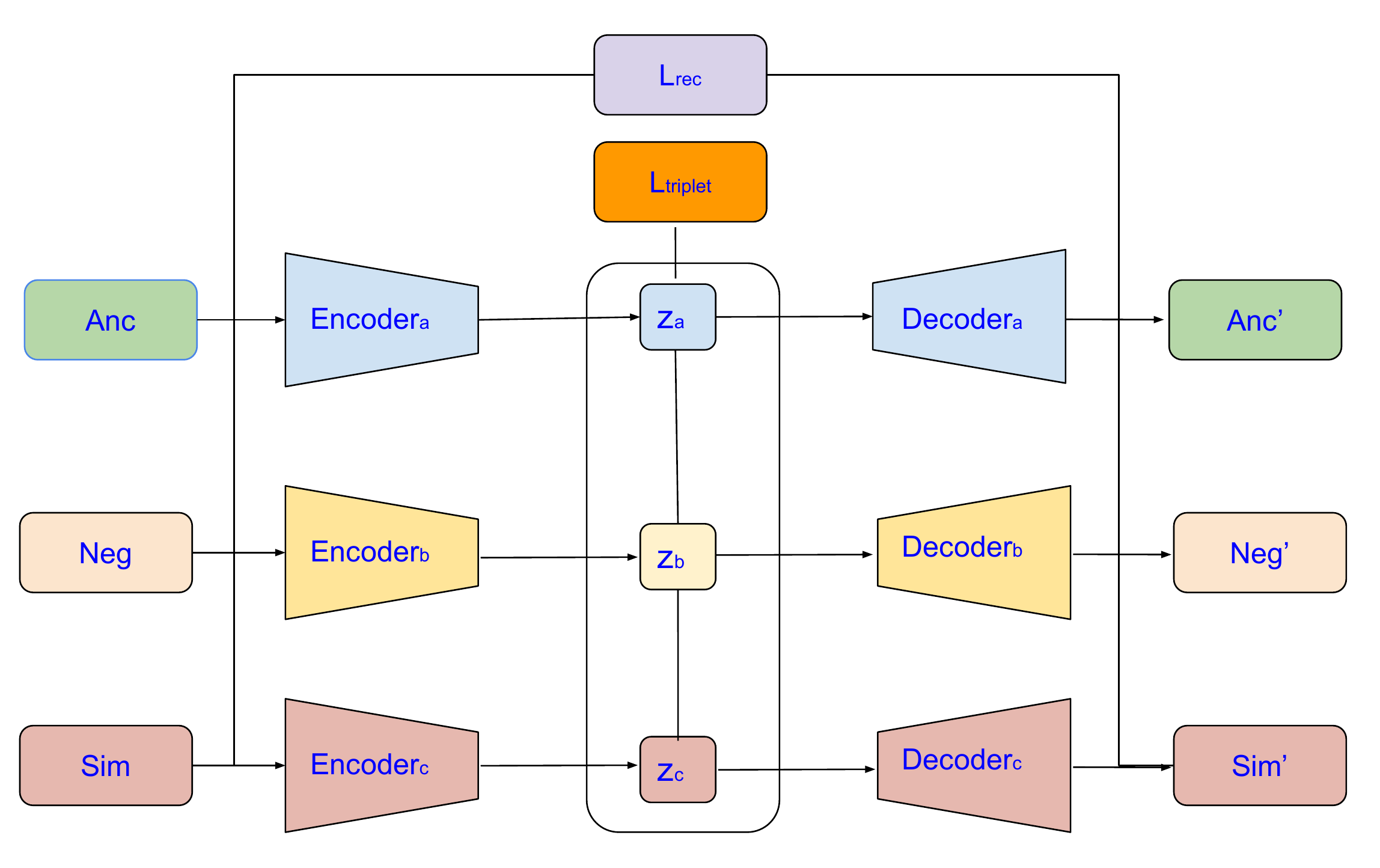}}
	\caption{$\rm{I^2CE}$ of triple branches.Compared to the dual branches structure, for this framework we introduce one more branch of anto-encoder to include similar sentences(Sim) and train with triplets.}
	\label{fig}
\end{figure}

\subsection{Training with Triplet Group Strategy}
For the semantic loss regularization, we also introduce a triplet training strategy. In the previous section, we add margin loss to make the model distinguish from sentences that have different meanings. Under the configuration of the triplet group training strategy, we add in neighboring sample pairs and form the (anchor, similar, negative) triplets for training. The elements of a triplet includes anchor, similar and negative sentences. Typically, the label sentences of the same group share some semantics since they describe the same image. Based on this consideration, we randomly select two sentences in the same group as an anchor and similar pair. And then we randomly choose one sentence from another group to form one triplet.

Triplet loss was proven useful in FaceNet\cite{schroff2015facenet} to learn distance aware face embedding, which means similar images would be pushed together and images with contrast differences located in distant area in the embedding space. The formula of triplet loss is shown as
\begin{equation}
{{\rm{L}}_{\rm{triplet}}} = \sum\limits_{i = 1}^N {{{[\alpha  + d(f(x_i^a),f(x_i^s)) - d(f(x_i^a),f(x_i^n))]}_ + }},
\end{equation}
where the N is the batch size, ($x_i^a$ , $x_i^s$ , $x_i^n$ ) is a grouped sentence triplet, the elements of which stand for anchor, similar and negative sentence, respectively. The function $f$ is to map a sentence to the learned embedding vector of a sentence. And the function $d$ is a kind of vector distance measure. This training objective would like the model to assign higher similarity scores(closer in semantic space) to the similar sample pair than to the not similar sample pair by a margin of at least alpha. Regularized by the semantic loss term the overall loss is expressed as
\begin{equation}
{\rm{L}_{\rm{overall}}} = {\lambda _1}{\rm{L}_{\rm{semantic}}} + {\lambda _2}{\rm{L}_{\rm{rec}}}.
\end{equation}
Here the $L_{semantic}$ can be the above defined $L_{margin}$ or $L_{triplet}$. The $\lambda _1$ and $\lambda _2$ are hyperparameters to balance the two kinds of losses.

\subsection{Cosine distance measure}
As mentioned above, for the semantic distance aware training objectives like triplet loss or margin loss, we need a distance metric for the sentence vectors. Some previous work use Manhatan distance\cite{mueller2016siamese} or Euclidean distance\cite{reimers2019sentence}. In this work we adopt cosine embedding distance. The formula for cosine distance is show as
\begin{equation}
{d_{\cos }}(({z_a},{z_b}),y) = \left\{ {\begin{array}{*{20}{c}}
	{1 - \cos ({z_a},{z_b}),}&{y = 1}\\
	{\max (0,\cos ({z_a},{z_b}) - \beta ),}&{y =  - 1}.
	\end{array}} \right.
\end{equation}
Here the $z_a$ and $z_b$ are the encoded vector representations of two sentences.After training, we use the representation given by the encoder and calculate the cosine distance between sentences with the indicator y equals to one.
\begin{equation}
{\rm{Sim}}(a,b) = 1 - {d_{\cos }}({z_a} , {z_b})
\end{equation}
Here, by computing the similarity between the intrinsic vectors $z_a$ and $z_b$ of (cand, ref) pair we got the similarity score $Sim(a,b)$ of the candidate caption. Typically there are five similarity scores considering that each of the label sentences may express different aspects of the image. In later section we will discuss on the strategy of scores pooling.

\section{Experiments}
In this section, we carry out several experiments to validate the effectiveness of the proposed $I^2CE$ method for evaluating the quality of image captions. Our motivation is to propose both intrinsic and semantic distance-aware sentence level embeddings for image caption evaluation.

\subsection{Implementation Details}
In the following section, we first introduce the dataset for training both the image captioning models and the proposed caption evaluation model $I^2CE$ and provide more training details for the evaluation metric.

Dataset. We use the label sentences from MS COCO dataset\cite{lin2014microsoft}. The numbers of paired images are (113287, 5000, 5000) for (training, validation, testing) respectively. Each image is paired with five label sentences, the meaning of which should be similar but may vary for some aspects.

Image Caption Models. In this paper we employ some of the current popular generation models including fc\cite{rennie2017self}, top-down\cite{anderson2018bottom}, att2in \cite{rennie2017self}, transformer\cite{vaswani2017attention} and with CIDER rewards optimization based on reinforcement learning.  

FC model only uses the FC features, Att2in and Transformer use the spatial attention features, and TopDown model uses both types of features. Each model is trained using a two-stage training strategy: 1) MLE with standard cross-entropy loss; 2) self-critical optimization with cider rewards\cite{rennie2017self}. For the caption metric evaluation part, we choose the att2in model as the captioning model. 

Image Caption Evaluation Metrics. In this paper we include the popular image captioning evaluation metrics including cider\cite{vedantam2015cider}, spice\cite{anderson2016spice},meteor\cite{banerjee2005meteor}, rouge\cite{lin2004rouge},bleu1-4\cite{papineni2002bleu}. Note that these contemporary adopted metrics are all rule based ones while our proposed metric is learning based.
To learn the proposed intrinsic caption evaluator $I^2CE$, we first build a GRU-based sentence auto-encoder to learn intrinsic sentence embeddings. And to make the learned embedding more semantic distance aware, we provide two options: margin loss and triplet loss. 

Training details. We set a margin of 0.2 as suggested by\cite{luo2018discriminability} for both margin and triplet training loss.  We use Adam optimizer \cite{kingma2014adam} to train the parameters of neural networks with an initial learning rate of 5e-4. We use batch size 128. The model is trained on a GPU of Titan 2080Ti.

\subsection{Experimental Results}
In this section, we conduct evaluation experiments to validate the effectiveness of the proposed $I^2CE$ metric. Note that our $I^2CE$ metric is learning-based, and the motivation is to provide semantic-aware sentence level embeddings to evaluate the image captions rather than n-gram level matching. 

Validation on $I^2CE$ method for semantic representation. We mainly adopt the caption level correlation with human judgments for the metric performance. We ask human annotators to score five hundred machine-generated captions from att2in model, given the paired images. The score for each caption is between zero and five.

Following previous studies on Image Captioning Evaluation\cite{vedantam2015cider, anderson2016spice, cui2018learning, jiang2019tiger, yi2020improving}, we use Kendall-Tau, Spearman-Rho and Pearson-Rho to calculate the caption level correlation between captioning metric scores and annotated scores collected from human judgments.

\begin{table}[]
	\caption{Performance Evaluation of human judgments correlation on $\rm{I^2CE}$ Score Pooling Strategy with $I^2CE\_contrast$}  % 
	\label{table}  % 
	\begin{tabular}{llllll}
		\hline
		& max1  & max2  & max3  & max4  & max5  \\ \hline
		kendall  & 32.72 & 34.3  & 34.54 & 32.96 & 30.85 \\
		spearman & 48.08 & 50.09 & 50.29 & 48.34 & 45.84 \\
		pearson  & 37.31 & 39.79 & 40.76 & 39.3  & 37    \\ \hline
	\end{tabular}
\end{table}

\begin{figure}[htbp]
	\centerline{\includegraphics[width=9.5cm]{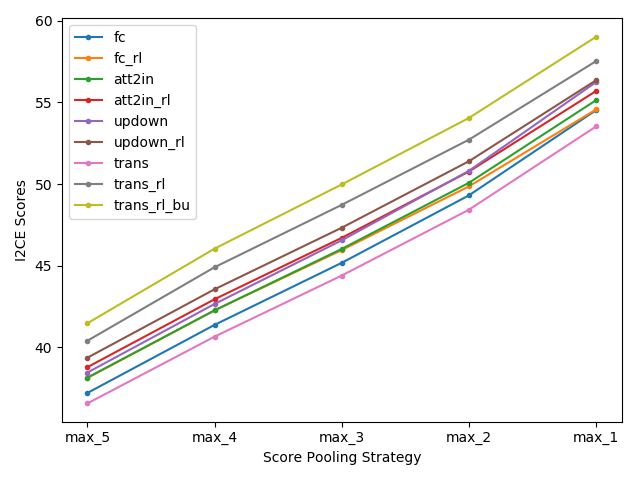}}
	\caption{$\rm{I^2CE}$ scores of state-of-the-art models on different score pooling strategies.}
	\label{fig}
\end{figure}

\textbf{Effects of Respected Scores Pooling Strategy.} We aim to learn sentence-level semantic representations while in the dataset for each image there are five label sentences. Generally, the contemporary metrics take the summary of matchings from all five annotations. Since each image has five sentences as ground truth which may vary and describes different aspects of the image, we calculate the similarities between the candidate caption and each of the label sentences. In experiments, we find that for each baseline of the $I^2CE$ model, combining more than one similarity score generally achieves better performance. For simplicity here, we only show the results of the dual branches approach. The best strategy is to take the average of the top three similarity scores among five ground truths. In later experiments, we set as default with this optimal strategy on pooling of scores. 

And we also select several caption model to test the effects taken by this score pooling strategy. The results are shown in Figure 4. The notation max-1 means taking the highest similarity score among five, max-2 means taking the average of the highest two similarity scores, and so on. As we can see, when more label sentences come in, the score decreases since generally, different labels may attend to different aspects of the image.

\textbf{Main results on metrics evaluation with ablation studies.} To validate the effectiveness of the proposed auto-encoding mechanism and distance-aware training strategy, we further evaluate the performance of $I^2CE$ on different configurations. 

\begin{table*}[]
	\caption{Performance Evaluation on $\rm{I^2CE}$ metric and contemporary automatic metrics}  % 
	\label{table}  % 
	\begin{tabular}{llllllllllllll}
		\hline
		& Human & SPICE & CIDER & $\rm{i2ce\_train}$ & $\rm{i2ce\_all}$ & $\rm{i2ce\_margin}$ & $\rm{i2ce\_triplet}$ & METEOR & ROUGE & B-1   & B-2   & B-3   & B-4    \\ \hline
		Kendall  & 39.11 & 30.15 & 32.36 & 31.38        & 32.21     & \textbf{34.54} & 32.56         & 29.27  & 28.46 & 30.58 & 29.65 & 25.82 & 21.42  \\
		Spearman & 62.12 & 44.93 & 47.38 & 46.73          & 47.14     & \textbf{50.29} & 48.02            & 43.68  & 42.87 & 47.01 & 44.46 & 21.03 & -18.95 \\
		Pearson  & 47.42 & 36.94 & 35.8  & 38.77       & 37.43     & \textbf{40.76} & 38.71         & 30.24  & 33.15 & 37.7  & 36.6  & 29.79 & 24.13  \\ \hline
	\end{tabular}
\end{table*}

\begin{table*}[htbp]
	\centering  % 显示位置为中间
	\caption{Caption Model Performance Evaluation using $\rm{I^2CE}$ metric and contemporary automatic metrics. B-1-4 stands for Bleu 1-4. C/10 stands for CIDER score normalized by 10. For all metrics the higher the better and the maximum value is one hundred.}  % 
	\label{table}  % 
	
	% l代表左对齐，c代表居中，r代表右对齐
	\begin{tabular}{|l|lllllllll|}
		\hline
		\textbf{Caption} &                            &                             &                             & \textbf{Caption}             & \textbf{Models}              &                                  &                            &                                &               \\ \cline{2-10} 
		\textbf{Metric}  & \multicolumn{1}{l|}{fc}    & \multicolumn{1}{l|}{fc\_rl} & \multicolumn{1}{l|}{att2in} & \multicolumn{1}{l|}{att2in\_rl} & \multicolumn{1}{l|}{topdown} & \multicolumn{1}{l|}{topdown\_rl} & \multicolumn{1}{l|}{trans} & \multicolumn{1}{l|}{trans\_rl} & trans\_rl\_bu \\ \hline
		B-1               & \multicolumn{1}{l|}{71.89} & \multicolumn{1}{l|}{75.31}  & \multicolumn{1}{l|}{74.06}  & \multicolumn{1}{l|}{76.83}      & \multicolumn{1}{l|}{75.5}    & \multicolumn{1}{l|}{78.28}       & \multicolumn{1}{l|}{70.66} & \multicolumn{1}{l|}{75.12}     & 78.38         \\ \hline
		B-2               & \multicolumn{1}{l|}{55.21} & \multicolumn{1}{l|}{58.25}  & \multicolumn{1}{l|}{57.39}  & \multicolumn{1}{l|}{60.11}      & \multicolumn{1}{l|}{59.2}    & \multicolumn{1}{l|}{61.86}       & \multicolumn{1}{l|}{52.94} & \multicolumn{1}{l|}{58.52}     & 62.42         \\ \hline
		B-3               & \multicolumn{1}{l|}{41.89} & \multicolumn{1}{l|}{43.08}  & \multicolumn{1}{l|}{43.35}  & \multicolumn{1}{l|}{45.31}      & \multicolumn{1}{l|}{45.23}   & \multicolumn{1}{l|}{47.07}       & \multicolumn{1}{l|}{39.33} & \multicolumn{1}{l|}{44.1}      & 47.99         \\ \hline
		B-4               & \multicolumn{1}{l|}{31.9}  & \multicolumn{1}{l|}{31.48}  & \multicolumn{1}{l|}{32.7}   & \multicolumn{1}{l|}{33.67}      & \multicolumn{1}{l|}{34.44}   & \multicolumn{1}{l|}{35.23}       & \multicolumn{1}{l|}{29.11} & \multicolumn{1}{l|}{32.69}     & 36.3          \\ \hline
		METEOR           & \multicolumn{1}{l|}{25.34} & \multicolumn{1}{l|}{25.3}   & \multicolumn{1}{l|}{25.8}   & \multicolumn{1}{l|}{26.42}      & \multicolumn{1}{l|}{26.75}   & \multicolumn{1}{l|}{27.07}       & \multicolumn{1}{l|}{24.87} & \multicolumn{1}{l|}{26.29}     & 27.76         \\ \hline
		ROUGE            & \multicolumn{1}{l|}{53.05} & \multicolumn{1}{l|}{53.99}  & \multicolumn{1}{l|}{54.15}  & \multicolumn{1}{l|}{55.55}      & \multicolumn{1}{l|}{55.56}   & \multicolumn{1}{l|}{56.44}       & \multicolumn{1}{l|}{51.71} & \multicolumn{1}{l|}{54.19}     & 56.71         \\ \hline
		C/10             & \multicolumn{1}{l|}{9.877} & \multicolumn{1}{l|}{10.58}  & \multicolumn{1}{l|}{10.2}   & \multicolumn{1}{l|}{11.19}      & \multicolumn{1}{l|}{10.87}   & \multicolumn{1}{l|}{11.83}       & \multicolumn{1}{l|}{9.45}  & \multicolumn{1}{l|}{10.91}     & 12.14         \\ \hline
		SPICE            & \multicolumn{1}{l|}{18.2}  & \multicolumn{1}{l|}{18.56}  & \multicolumn{1}{l|}{19.04}  & \multicolumn{1}{l|}{19.75}      & \multicolumn{1}{l|}{20.2}    & \multicolumn{1}{l|}{20.53}       & \multicolumn{1}{l|}{18.75} & \multicolumn{1}{l|}{20.42}     & 21.63         \\ \hline
		WMD              & \multicolumn{1}{l|}{21.88} & \multicolumn{1}{l|}{21.56}  & \multicolumn{1}{l|}{21.98}  & \multicolumn{1}{l|}{22.95}      & \multicolumn{1}{l|}{23.6}    & \multicolumn{1}{l|}{23.98}       & \multicolumn{1}{l|}{21.19} & \multicolumn{1}{l|}{22.93}     & 25.35         \\ \hline
		$\rm{I^2CE}$             & \multicolumn{1}{l|}{45.18} & \multicolumn{1}{l|}{45.96}  & \multicolumn{1}{l|}{46.03}  & \multicolumn{1}{l|}{46.71}       & \multicolumn{1}{l|}{46.56}   & \multicolumn{1}{l|}{47.33}       & \multicolumn{1}{l|}{44.4} & \multicolumn{1}{l|}{46.7}     & 48.19         \\ \hline
	\end{tabular}
\end{table*}

We provide four different settings of $\rm{I^2CE}$: 1) $\rm{I^2CE}$+train split, which only uses the reconstruction loss training on label sentences from MS COCO training splits; 2) $\rm{I^2CE}$+full splits, which uses all labels sentences of MS COCO; 3) $\rm{I^2CE}$+full splits+triplet loss, which uses both reconstruction loss and triplet loss training on the full corpus of MS COCO; 4) $\rm{I^2CE}$+full splits + margin loss, which uses margin loss on negative sentence pairs to guide the model to learn the different semantics of sentences with different meanings. Together with other contemporary captioning metrics, we test the correlation between the captioning metric scores and human judgments. The main results are presented in the middle part of following Table 1.

When training without any explicit semantic distance aware training objectives, the $\rm{I^2CE}$ metric can achieve a close performance to CIDER. To our knowledge, this should own to the effects of self-supervised learning embedded in the auto-encoding mechanism. By reconstructing the original sentences, the model can learn to abstract the most important information of the input sentence, which shows great potentials for applying auto-encoder to representation learning.And by adding more training corpus the performance is slightly better except for the Pearson correlation.
With margin loss, the performance boosts with a large margin comparing to the original baseline. Guided with an explicit distance-aware training objective which may serve as a regularization term on the original reconstruction loss, the $I^2CE\_contrast$ model achieves better performance to cider and spice.

\textbf{Reasons for margin loss improving human judgments correlation.} Results in Table 1 show that the two branches model trained on entire corpus obtain the best performance. Firstly this should own to the five annotations of each image. Each annotation points to some attended aspects of the paired image, which provides different semantic information for the candidate caption to match. We also notice that the three branches approach did not bring as good performance-boosting as to the two branches approach. We think this is mainly because selecting two sentences from the five annotations of the paired image may not be proper to learn similar semantic information. The annotators may attend to different parts of the image when they give the descriptions. Thus the meaning of annotations of the same group may vary. However, randomly selecting two sentences from different paired images can provide a sentence pair with different meanings, which forms the negative pair for margin loss.

\textbf{Comparing with Contemporary Metrics.} Results in Table1 show that the $I^2CE$ metric trained with both reconstruction loss and margin loss obtains better performance on human judgments correlation among all baselines. That experiment keeps the captioning model fixed with att2in model. To further validate the effectiveness of the proposed $I^2CE$ method, we add more captioning models to test their performance on different metrics. The results are shown in Table 3. We can see that attentive models like att2in and topdown model outperform FC model, which only use fully connected layer features. And models trained with CIDER rewards optimization generally achieve better results. The transformer model trained with bottom-up features and CIDER rewards achieves the best performance on all metrics. The $I^2CE$ metric aligns well with other contemporary metrics on distinguishing among captioning models with different performances. An interesting phenomenon is that the $I^2CE$ metric shows especially good alignments to the rouge metric, which may indicate that the $I^2CE$ metric can be applied to text summary.

\begin{figure}[htbp]
	\centerline{\includegraphics[width=9.5cm]{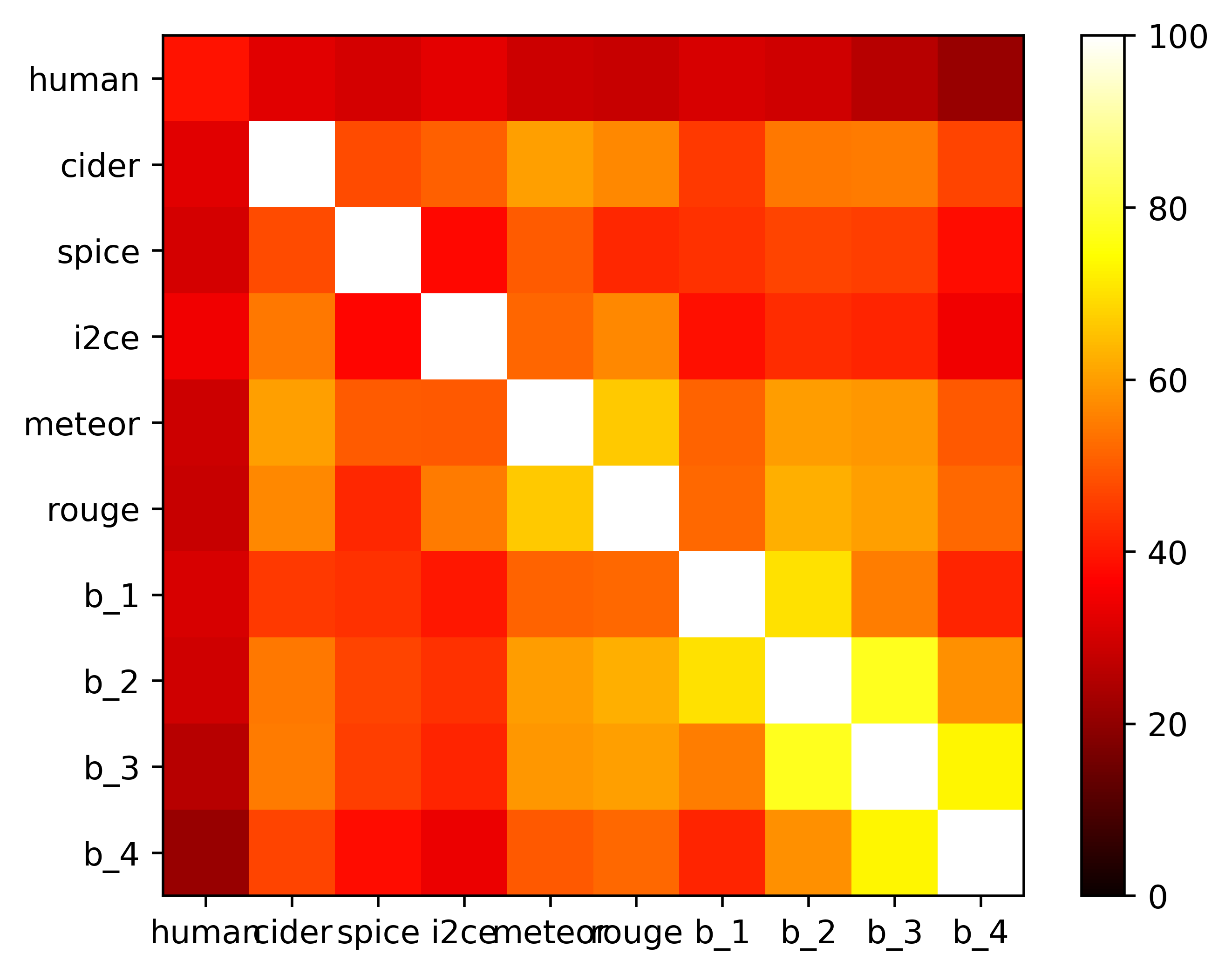}}
	\caption{Kenall-Tau correlation among metrics}
	\label{fig}
\end{figure}

\begin{figure}[htbp]
	\centerline{\includegraphics[width=9.5cm]{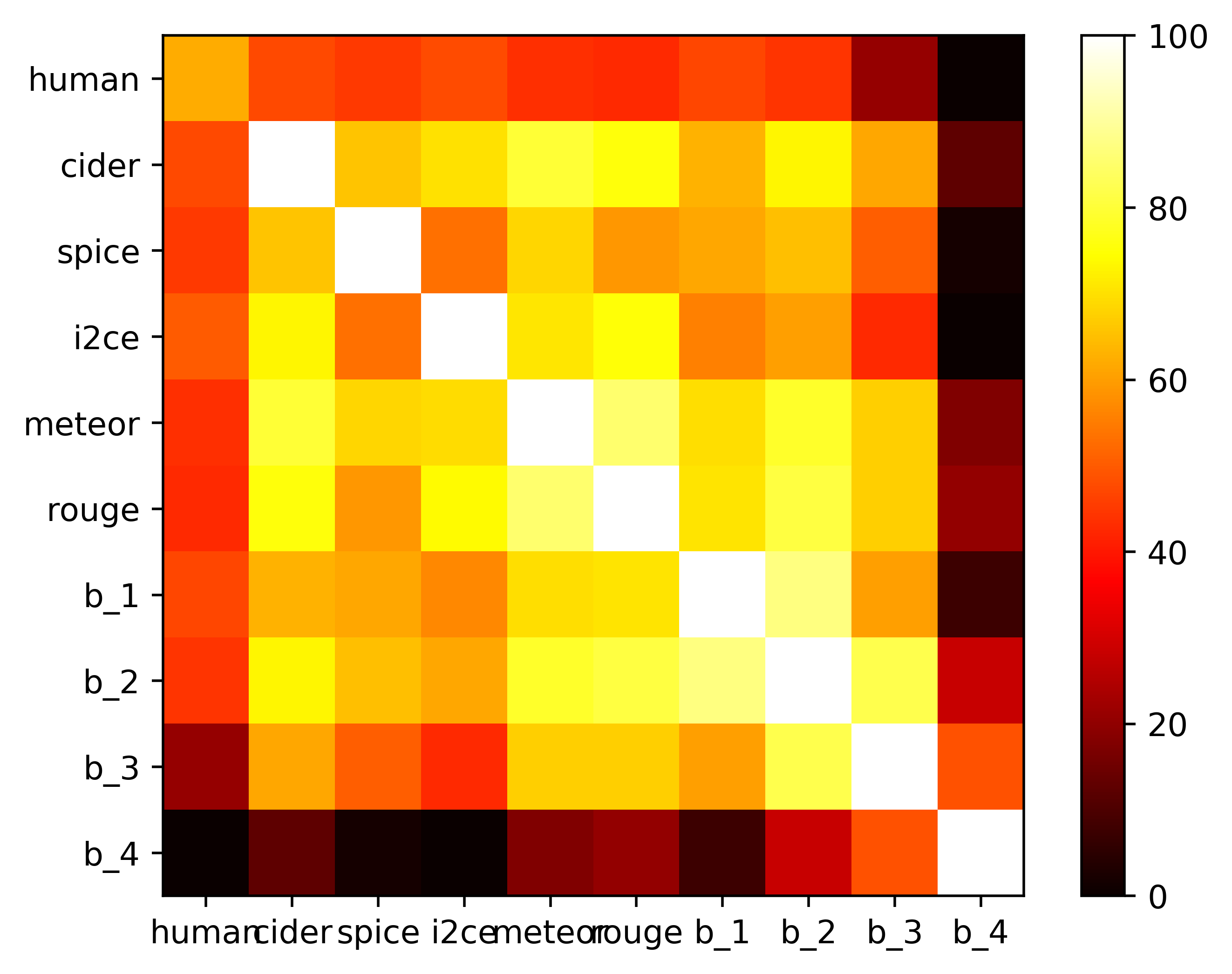}}
	\caption{Spearman-Rho correlation among metrics}
	\label{fig}
\end{figure}

\begin{figure}[htbp]
	\centerline{\includegraphics[width=9.5cm]{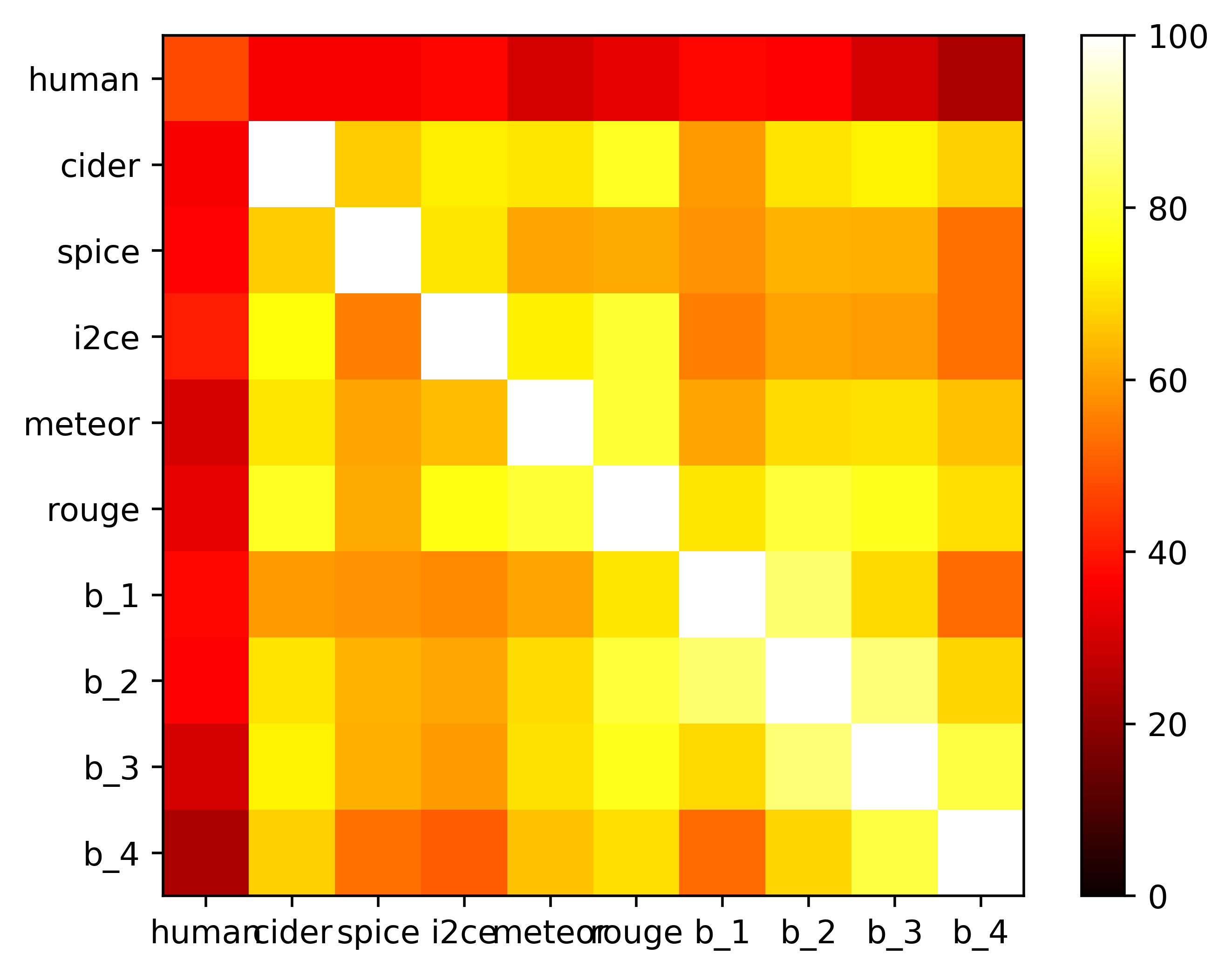}}
	\caption{Pearson-Rho correlation among metrics}
	\label{fig}
\end{figure}

\textbf{More results for metrics correlation.}  In Figure 5-7, we show the correlation matrix between any two captioning metrics tested, including human judgments. The results show that our $\rm{I^2CE}$ model also correlates well to the contemporary metrics, especially for rouge and cider. 

During the test, we observe that people score the candidate captions largely depending on the key visual words. For example, if a visual word in the candidate caption did not appear on the pairing image, which is a mistake made by the caption model, the quality score will be reduced by a large margin. And if the visual words mentioned in the candidate are all correct, then the score results are generally good.

However, since the 4-gram matching generally is hard to achieve for the caption model. Thus, the accurate 4-gram matches could be quite sparse, which could make the correlation value be negative, as shown in the Spearman-Rho measure.

\subsection{Intuitive Results on $\rm{I^2CE}$ Scores}
The following Fig.8 shows us the scores given by $\rm{I^2CE}$ on some testing cases with the testing captioning model $att2in+$. In the figure, there are three pieces of testing cases shown. For each testing case, there are one candidate caption in the first line and the five references below. Addicted to each reference is the respected $\rm{I^2CE}$ score for each (candidate, reference) pair. The experiment results of previous sections show that the best pooling strategy is to take the average of the top three similarity scores among the five label sentences. We denote the final $\rm{I^2CE}$ score in a seperate column.

\begin{figure}[htbp]
	\centerline{\includegraphics[width=9.5cm]{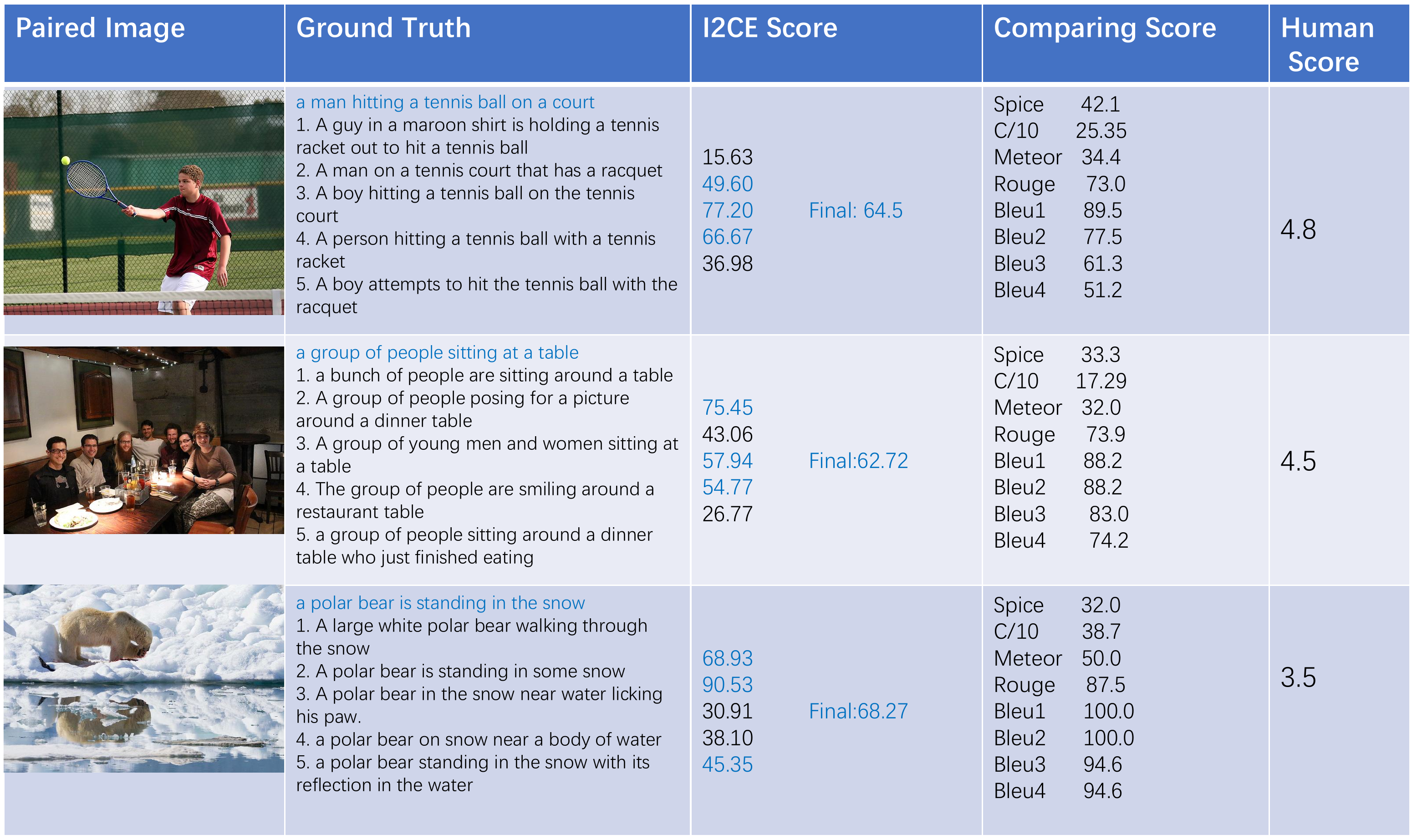}}
	\caption{$\rm{I^2CE}$ scores on highly aligned generated captions.}
	\label{fig}
\end{figure}

From the first picture in Figure 9, we can see that for the middle three generated captions which are most similar references given the candidate caption, get relatively higher scores. For the first reference, it has some additional details like "in a maroon shirt" and "with a racket", which are not covered by the candidate caption, and thus the similarity score is lower. The last reference is quite likely the same case. On contrast, the middle three references are relatively expressing the closer information as the test caption with fewer variances on details. Thus the candidate would reach higher scores respectively.

\begin{figure}[htbp]
	\centerline{\includegraphics[width=9.5cm]{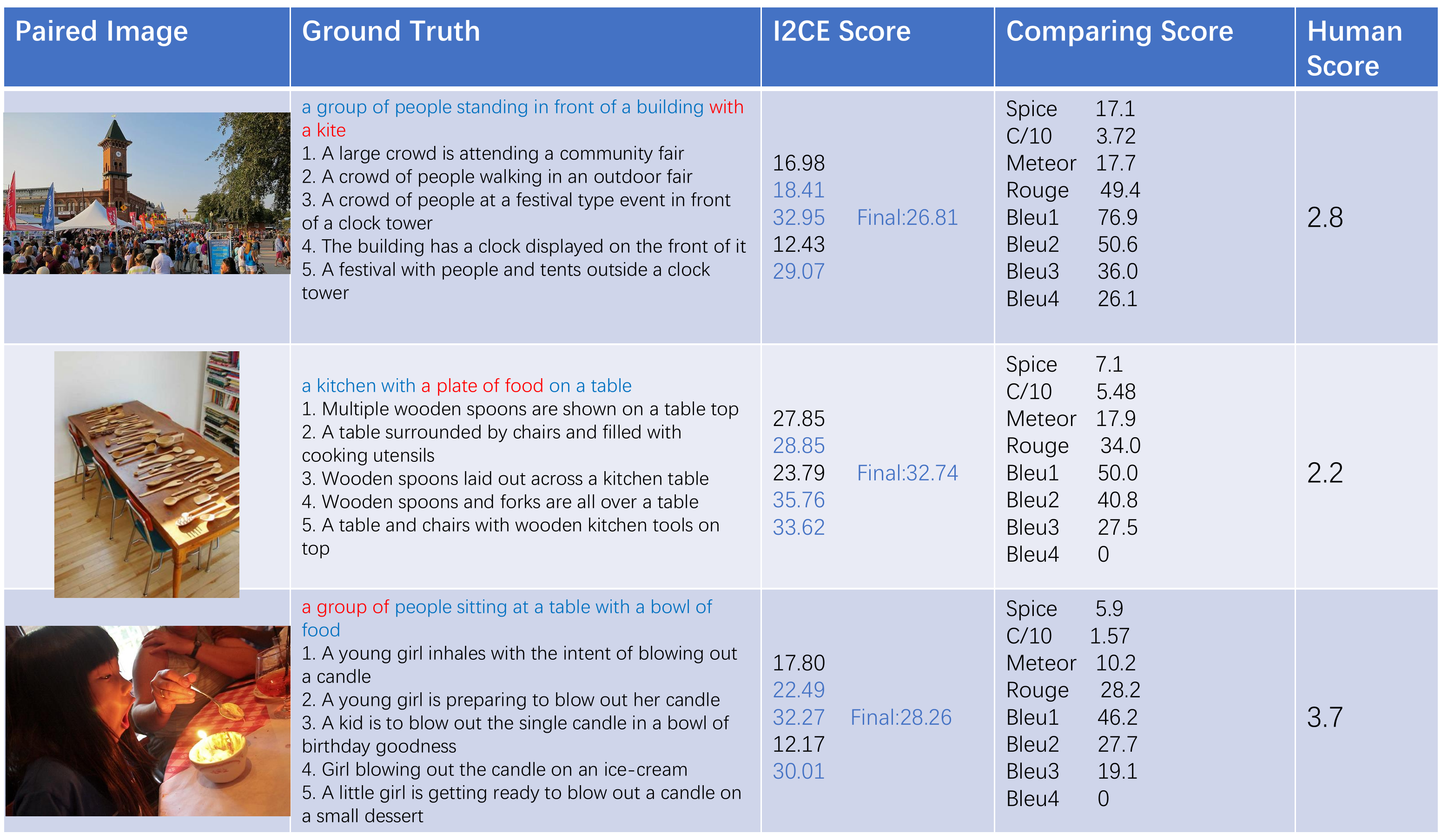}}
	\caption{$\rm{I^2CE}$ scores on less aligned generated captions.}
	\label{fig}
\end{figure}

We present some less aligned cases given in Fig.10. For these captions less aligned to the references, there are whether miss detected conceptions like "with kite" and "a plate of food", or syntactic sentence structure variances. For the third picture the candidate description seems to be right. However the main theme of the picture is that the little girl blowing out candle, which is not mentioned in the candidate caption. Thus all metric scores are relatively lower.

\section{Conclusion}
In this paper, we introduce the proposed $\rm{I^2CE}$ metric, which uses the intrinsic sentence vectors to calculate similarity instead of matching on n-gram tokens or word chunks. $\rm{I^2CE}$ benefits from sentence intrinsic information and gains understanding of semantic similarity in sentence level.  To make the intrinsic vectors more distance aware, we further develop two different varies of the original single branch approach which provides with explicit distance loss.

To validate the effectiveness of the proposed $I^2CE$ method, we conduct several experiments for evaluation. First, we evaluate all captioning metrics on human judgments correlation, in which $I^2CE$ with margin loss achieves the best performance comparing to contemporary metrics. We also apply all the caption metrics to several state-of-the-art captioning models and compare all the metric scores. $\rm{I^2CE}$ aligns with other metrics well and shows its effectiveness in distinguishing between captioning models with different performances. In addition, we also present some of the intuitive results on caption quality scores calculated by the $I^2CE$ and other metrics. The results show that $\rm{I^2CE}$ can attend more flexibly to the semantic meaning of candidate captions.
The $\rm{I^2CE}$ metric attends to an intrinsic semantic approach on evaluating similarity in sentence-level, which is not covered by the contemporary metrics. It may
serve as a complementary metric to give an intrinsic-oriented evaluation on candidate captions.Future work may involve with learning visual and textual joint embedding for image captioning evaluation.

\section*{Acknowledgment}

This work was supported by grants from the Research Grants Council of the Hong Kong Special Administrative Region, China, and from the City University of Hong Kong.

\bibliographystyle{IEEEtran}
\bibliography{refs.bib}

\end{document}